\documentclass[conference]{IEEEtran}

\IEEEoverridecommandlockouts

\usepackage{cite}
\usepackage{amsmath,amssymb,amsfonts}
\usepackage{threeparttable}

\newcommand{\linebreakand}{%
  \end{@IEEEauthorhalign}
  \hfill\mbox{}\par
  \mbox{}\hfill\begin{@IEEEauthorhalign}
}
\makeatother 

\usepackage{graphicx}
\usepackage{textcomp}
\usepackage[dvipsnames]{xcolor}
\usepackage{subfigure}
\usepackage{subcaption}
\usepackage{enumerate}
\usepackage{algorithm}
\usepackage{float}
\usepackage[noend]{algpseudocode}
\usepackage{diagbox}
\usepackage{booktabs}
\usepackage{adjustbox}
\usepackage{comment}
\usepackage{multirow}
\usepackage{pifont}
\usepackage{hyperref}

\DeclareRobustCommand{\[}{\begin{equation}}
\DeclareRobustCommand{\]}{\end{equation}}

\def\BibTeX{{\rm B\kern-.05em{\sc i\kern-.025em b}\kern-.08em
    T\kern-.1667em\lower.7ex\hbox{E}\kern-.125emX}}

\usepackage[final,commandnameprefix=always]{changes}
\definechangesauthor[color=BrickRed,name=Tozammel]{TH}

\begin{document}

\title{A Novel Multi-layer Task-centric and Data Quality Framework for Autonomous Driving}

\author{\IEEEauthorblockN{Yuhan Zhou}
\IEEEauthorblockA{\textit{Dept. of Information Science} \\
\textit{University of North Texas}\\
Denton, Texas, USA \\
yuhanzhou@my.unt.edu}

\and
\IEEEauthorblockN{Haihua Chen}
\IEEEauthorblockA{\textit{Dept. of Data Science} \\
\textit{University of North Texas}\\
Denton, Texas, USA \\
Haihua.Chen@unt.edu}
\and
\IEEEauthorblockN{Kewei Sha}
\IEEEauthorblockA{\textit{Dept. of Data Science} \\
\textit{University of North Texas}\\
Denton, Texas, USA \\
Kewei.Sha@unt.edu}
}

\maketitle

\thispagestyle{plain}
\pagestyle{plain}

\begin{abstract} 
The next-generation autonomous vehicles (AVs), embedded with frequent real-time decision-making, will rely heavily on a large volume of multisource and multimodal data. In real-world settings, the data quality (DQ) of different sources and modalities usually varies due to unexpected environmental factors or sensor issues. However, both researchers and practitioners in the AV field overwhelmingly concentrate on models/algorithms while undervaluing the DQ. To fulfill the needs of the next-generation AVs with guarantees of functionality, efficiency, and trustworthiness, this paper proposes a novel task-centric and data quality vase framework which consists of five layers: data layer, DQ layer, task layer, application layer, and goal layer. The proposed framework aims to mapping DQ with task requirements and performance goals. To illustrate, a case study investigating redundancy on the nuScenes dataset proves that partially removing redundancy on multisource image data could improve YOLOv8 object detection task performance. Analysis on multimodal data of image and LiDAR further presents existing redundancy DQ issues. This paper opens up a range of critical but unexplored challenges at the intersection of DQ, task orchestration, and performance-oriented system development in AVs. It is expected to guide the AV community toward building more adaptive, explainable, and resilient AVs that respond intelligently to dynamic environments and heterogeneous data streams. Code, data, and implementation details are publicly available at: \url{https://anonymous.4open.science/r/dq4av-framework/README.md}. 
\end{abstract}


\section{Introduction}

The recent deployment of RoboTaxi by Waymo, Tesla, and Mobileye demonstrates the transformative potential of autonomous driving (AD) to improve transportation safety, reduce traffic congestion, and enhance mobility for all \cite{omeiza2021explanations, guo2024advances, liuautonomous}. However, the widespread adoption of autonomous vehicles (AVs) still faces significant challenges, particularly in ensuring real-time, accurate, and reliable decision-making in a dynamic environment that involves long-tail risks and high uncertainty \cite{zhang2024multimodal, han2023collaborative, Chen2024E2E, guo2024advances, Liang_2024_AIDE, li2023open}. These challenges are compounded by constrained computational and storage resources in AVs. 

AVs make intelligent decisions using machine learning (ML) models trained on multisource, multimodal data collected from diverse sensors. The efficiency and effectiveness of AV decisions depend not only on the capacity of ML models but also critically on the quality of data \cite{chen2021data, sambasivan2021everyone,garbage2020}. While significant progress has been made in model development and benchmark-driven evaluation, current AV frameworks often overlook the critical role of data quality (DQ) in determining system performance, such as accuracy, robustness, and efficiency \cite{li2024datacentricAV, li2023open, li2022traffic, goknil2023systematic}. 

Existing data quality frameworks \cite{rangineni2023analysis, Priestley2023, schwabe2024metric, rahal2025enhancing} are mostly designed for traditional, homogeneous datasets and fall short in addressing the following complex demands of AV systems. First, AVs must handle multisource, multimodal data streams while simultaneously supporting a wide range of applications, each of which may impose different quality requirements on shared or unique data streams. Second, cost constraints resources in AVs, making it essential to carefully balance data quality with task performance. Third, AVs operate under stringent real-time constraints, where even a correct decision is ineffective if it arrives too late. Thus, a mechanism to evaluate the data quality and select an appropriate subset of data that satisfies all the above requirements is vital. 

While prior research has explored dataset characterization \cite{guo2024advances, li2024datacentricAV, li2023open, liu2024DatasetSurvey}, curation \cite{sun2020scalability, caesar2020nuscenes, geiger2013vision, alibeigi2023zenseact}, generation \cite{yang2020surfelgan, chen2024driving, xu2024drivegpt4}, and testing \cite{guo2024advances, Chen2024E2E}, to the best of our knowledge, there remains a lack of a unified framework that systematically addresses data quality in the AV context. To bridge this gap, we take a first step to propose a task-centric, multisource, and multimodal data quality framework tailored for AVs. This five-layer framework uses decision-making tasks as the central link between data quality metrics and AV performance objectives, offering a novel perspective for evaluating and improving data quality in real-time, safety-critical scenarios. To demonstrate the effectiveness of the task-centric framework, we use redundancy as a representative data quality metric and the object detection as the target task. The evaluation results show that redundancy exists in multisource and multimodal data, and removing redundancy could have little impact on or even improve the task performance. This evidence emphasizes the significance of data quality and the necessity of a framework for evaluating DQ under a specific AV context.

The contributions of this paper can be summarized as follows. 

\begin{enumerate}
    \item \textbf{A five-layer Vase Framework with tightly coupled layers—Data, Data Quality, Task, Application, and Goal.} Our system treats driving tasks as first-class entities, orchestrates data pipelines in real-time, and evaluates decisions through multidimensional data quality indicators. We not only highlight these critical yet overlooked factors but also advocate for recognizing the relationships among them for iterative improvement. 
    \item \textbf{An empirical evaluation to validate the proposed framework.} Implemented on nuScenes, we removed the redundant instances among the multisource images from six cameras. Use pruned datasets to train the object detection model YOLO \cite{terven2023yolo} could achieve the same or even higher task performance. Experiments on redundant multimodal data of image and LiDAR also prove the significance of task-centric data quality evaluation.   
    \item \textbf{Open discussions and future directions.} We outline a few critical research questions and directions under the multi-layer task-centric and data quality framework. These open research questions will inspire and guide the community to explore more efficient and reliable techniques for AV.
\end{enumerate}

The rest of the paper is organized as follows. In Section \ref{sec: framework}, we introduce and explain the Vase Framework, each layer, relationships, and the rationales behind the design. Under this, we conduct experiments in Section \ref{sec: case} on redundancy to illustrate the data quality evaluation in the framework using the nuScenes dataset, which consists of image and LiDAR data for object detection. At last, in Section \ref{sec: future} we put forward open questions to discuss, and summarize the future directions of the data quality work in AV. We conclude the paper in Section \ref{sec:con}.

\section{A Novel Task-centric and Data Quality Vase Framework for AVs}
\label{sec: framework}

Traditional frameworks present the following research gaps. (1) They mainly emphasize model-centric architectural designs and overlook data quality despite its significance in assuring the performance, efficiency, and reliability of AVs \cite{li2024datacentricAV, li2023open, kim2025automatic}. (2) The quality of multisource and multimodal data in AV varies under dynamic environments \cite{sun2020scalability, caesar2020nuscenes, geiger2013vision, alibeigi2023zenseact, zhang2024multimodal}. Existing DQ frameworks are not well-suited to handle these complexities, highlighting the need for a novel DQ framework tailored to the unique challenges of AV systems. (3) Current strategies not only overlook DQ evaluation but also treat all the AV tasks with similar DQ requirements \cite{li2024datacentricAV}. Consequently, the lack of DQ evaluation and task-centric awareness prevents AV systems from feeding quality diagnostics back into the loop and iterating the performance.

\begin{figure*}[ht]
    \centering
    \includegraphics[width=1\linewidth]{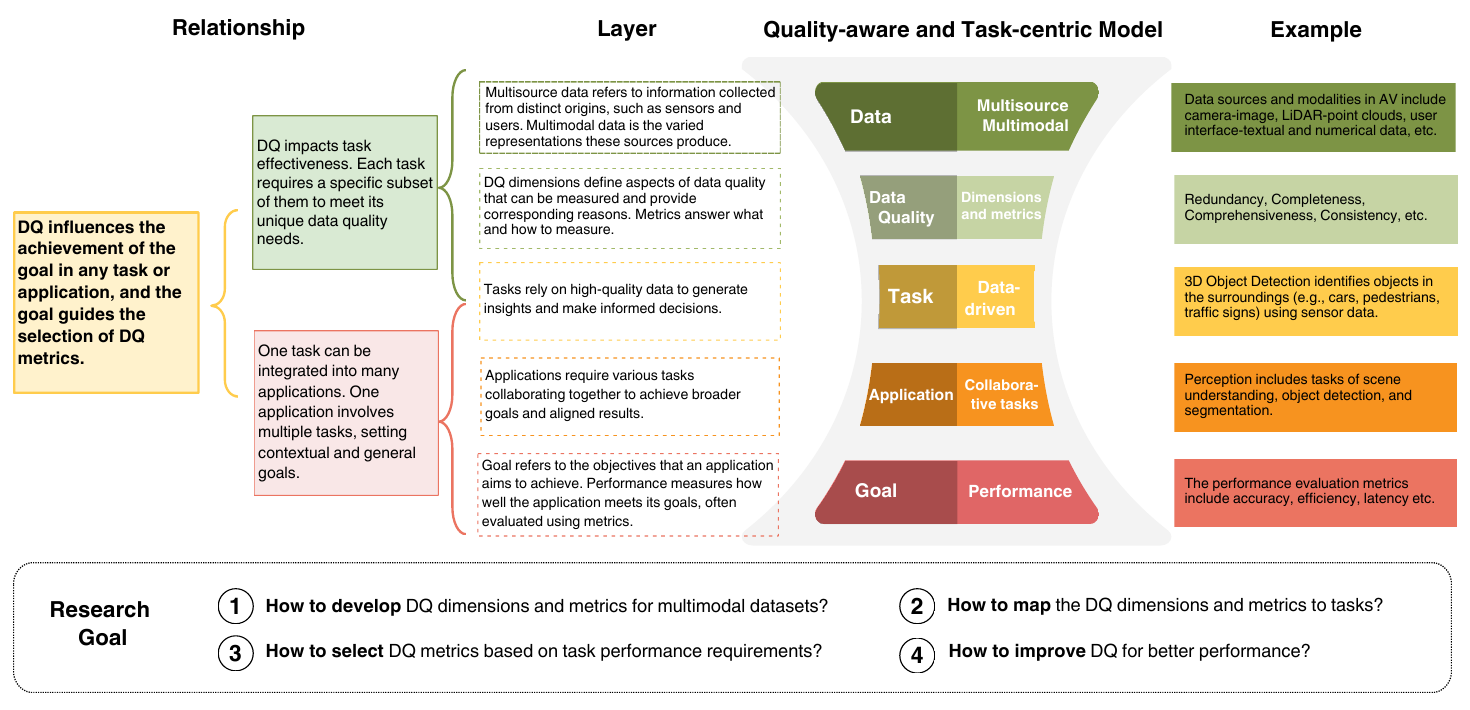}
    \caption{The multi-layer task-centric and data quality Vase Framework}
    \label{fig: framework}
\end{figure*}

Motivated by addressing above gaps and inspired by the five-layer architecture of the Internet of Things \cite{wu2010IoTLayer}, we propose a multi-layer task-centric and data quality Vase Framework in Fig. \ref{fig: framework}. It integrates five layers, with corresponding examples, explanations, and relationships. Adopting AD as the context, it begins with the data collected from multiple sources in various modalities. Data quality, task, and goal layers are central to this structure. Each layer is introduced in Section \ref{subsec: layer}, and the relationships are discussed in Section \ref{subsec: relationships}.

Research goals at the bottom highlight the significance of our work. We aim to address these in future work through extensive studies. These research questions are critical because they link data quality and the performance of AV systems. They guide us in defining the DQ dimensions that directly affect key tasks and selecting a specific data quality set based on task requirements. By addressing these questions, we hope to develop more robust, adaptable, and safe AV systems. More detailed questions are listed and discussed in Section \ref{sec: future}.

\subsection{Five-Layer Framework Breakdown}
\label{subsec: layer}

\begin{figure}
    \centering
    \includegraphics[width=1\linewidth]{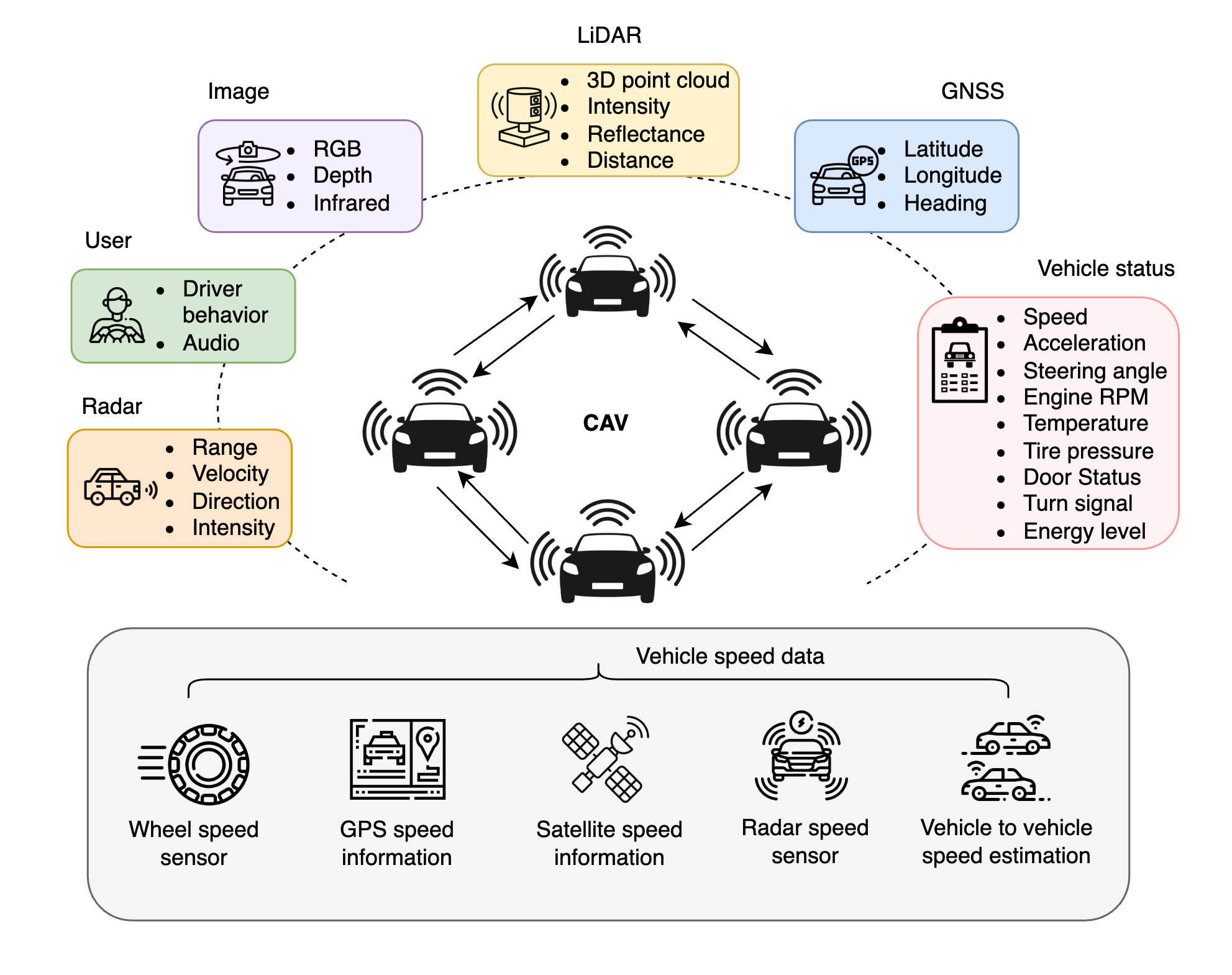}
    \caption{Illustration of multisource and multimodal data in autonomous driving}
    \label{fig: multi-source}
\end{figure}

\paragraph{Data Layer}
Data in AVs presents two main features: multisource and multimodal, as shown in Fig. \ref{fig: multi-source}. At the \textbf{multisource} level, AVs gather information from various sensors, such as LiDAR, cameras, and radar, as well as other feeds like information collected from connected autonomous vehicles (CAVs), user behavior data, weather reports, and infrastructure updates. Since one AV can be equipped with more than one camera, the front, back, and side cameras are considered different data sources \cite{zhou2020end}. This blend of inputs can validate data and reduce the impact of errors or noises from individual sensors, which enables a more comprehensive view of the driving environment and a more accurate and robust performance \cite{zhang2022sensor, song2019apollocar3d}. 

Modalities are the corresponding data types generated by the sources and widely discussed in data fusion techniques \cite{fu2024eliminating, zhao2024deep}. \textbf{Multimodal} in AV refers to the diverse forms and types in which data is captured and represented, including raw sensor readings, images, textual data, video, and audio \cite{yang2024deepinteraction++}.  

Combining multisource and multimodal data can enhance the performance and reliability in AV compared to a single data source or modality. Zhang et al. improved vehicle localization by integrating vehicle headlights and taillights detected from multiple cameras \cite{zhang2022night}. Sanchez et al. found that single-source LiDAR detectors generalize poorly when AVs enter unfamiliar zones, and multisource LiDAR training can improve domain generalization by over 10\% and source-to-source transfer accuracy by 5.3\% over any individual dataset \cite{sanchez2025cola}. Sensor complementarity is another reason why multisource and multimodal data are essential. Under night, fog, rain, or snow, camera performance degrades, and fusing radar or LiDAR restores robustness, increasing detection score by up to 29\% \cite{su2024adverse}. In dense urban scenarios, cameras often miss vehicles occluded by large objects, while LiDAR and radar reveal them \cite{hou2023fault}. When LiDAR points are sparse for tiny objects, a camera could help recover them by visual cues \cite{fu2024eliminating, wang2023scenes, wang2023od, yang2024deepinteraction++}.

\paragraph{Data Quality Layer}

While multisource and multimodal data enhance AV perception and decision-making, they also introduce data quality challenges with varying accuracy and noise. Conflicting information from various sources and modalities could impact data consistency and correctness. Fusing these data may cause calibration incorrectness,  redundant information \cite{zhu2023survey}, and imbalanced issues \cite{zhang2024LowQuality}. To address these, a structured data quality layer can contribute to the evaluation and improvement of data. The following introduces key data quality dimensions designed for multisource and multimodal data in AV to emphasize the quality-aware focus of this framework. Table \ref{table: DQ} presents the main and common DQ dimensions, which could be expanded or tailored based on different scenarios.

\begin{table*}[htbp]
  \centering
  \caption{Data quality dimensions, explanations and their impacts in AV tasks}
  \label{table: DQ}
  \begin{tabular}{%
    p{1.8cm}  
    p{5.5cm}  
    p{3.0cm}  
    p{6.5cm}  
  }
    \toprule
    \textbf{DQ dimension} &
    \textbf{Explanation} &
    \textbf{AV task} &
    \textbf{Impact} \\
    \midrule
    Completeness &
    The extent to which all required data elements are present for a task. For labeling, it ensures all valid objects in a frame are labeled \cite{emmanuel2021survey, chen2021data}. &
    Object detection and tracking &
    Missing annotations and bounding boxes reduce recall and increase bias \cite{kim2025automatic}. \\

    Consistency &
    The extent to which information provided by sensors and modalities is non-contradictory \cite{li2024light, rao2020quality, pradhan2024identifying, kong2025multi, fu2024eliminating}. &
    Vehicle recognition and labeling &
    Cross-modal divergence can trigger a high-fault flag in an information-fusion monitor ~\cite{hou2023fault}. \\

    & Data remains non-contradictory over time without discrepancies due to latency, environmental changes, or synchronization issues \cite{you2021temporal}. &
    Semantic perception &
    The representation and labels for the same object viewed at different points in time should be the same \cite{Nunes_2023_Temporal}.\\

    Correctness & The extent of data values or labels reflects real-world entities, positions, and attributes \cite{chen2021data, kim2025automatic, pradhan2024identifying}. &	
    Object detection under adverse weather &	
    Raindrops and snow on windshields or camera lenses impair visibility \cite{luo2024impact, brophy2023review}. \\

    Noise level	& The degree to which a sensor stream is contaminated by random or systematic distortions that obscure the true signal.&	
    Object detection for night-time urban driving \cite{zhang2022night} &
    High‐ISO noise and motion blur in camera images inflate the false-negative rate.\\

    Redundancy & 
    The presence of overlapping information does not increase task performance \cite{zhang2022sensor, chen2023survey, li2023reducing}.	&
    Scene understanding	&
    Consecutive LiDAR data are highly redundant due to the high sampling frequency and the massive surrounding environment information besides the objects of interest \cite{yuan2021temporal, zhang2024igo}. \\

    Relevance &
    The degree to which the data contributes useful information to the current task. & 
    Trajectory prediction and planning & 
    Irrelevant or weakly related inputs can cause unsafe actions, such as reacting to distant objects first. \\

    Timeliness	& Data is delivered and processed before a deadline \cite{guo2023AoI}. & 	Prediction and path planning & If the trajectory prediction of localized objects misses its deadline, the planner may make outdated motion hypotheses, leading to unsafe conditions \cite{Shi2024timeliness}.\\

    \bottomrule
  \end{tabular}
\end{table*}

Multisource and multimodal data in AVs introduce unique challenges that necessitate a more rigorous data quality evaluation framework \cite{zhang2024LowQuality}. DQ dimensions of multisource and multimodal data in AVs are similar but not identical. Both types require correctness, consistency, relevance, and timeliness to ensure reliable perception and decision-making. At the same time, the definition of the same DQ dimension may vary for multisource and multimodal data. For example, in multimodal data, \textbf{completeness} refers to capturing and integrating all necessary data from different sensors, whereas in multisource data, it emphasizes collecting data that covers the required spatial areas and time intervals. \textbf{Redundancy} can arise from duplicate detections by multiple sensors on the same vehicle \cite{sun2020advanced} or from overlapping reports across AVs and infrastructure nodes. Managing redundancy effectively prevents unnecessary data processing while preserving critical information for robust task performance. Redundant multimodal data occurs when multiple sensors provide overlapping information, which can be beneficial for fault tolerance and safety \cite{liu2021Multi-sensorRedundancy, qian20223d} but inefficient if excessive \cite{Zhao_2023_ada3d, li2023reducing, Sun2023Task-Driven, xu2024dm3d}.

\paragraph{Task Layer}

Data quality evaluation and improvement should follow the principle of ``fit for purpose" \cite{wang1996}. There are two reasons for the \textbf{task-centric} layer design: (1) The specific task requirements should guide the selection of data quality dimensions. Each task also involves various conditions, such as weather, scenarios, and light, impacting its own set of critical quality attributes. (2) In real-world scenarios, gathering perfect data to meet all quality dimensions is challenging. Therefore, it is essential to select the most critical dimensions that fit specific tasks and contexts in AV. 

Targeted data quality improvements can directly enhance the performance of each module in the closed-loop system. Tasks such as object detection, segmentation, and tracking require investigation and comparison of the detailed characteristics of the dataset \cite{liu2024DatasetSurvey}. For instance, tasks that involve real-time decision making, such as obstacle detection, may prioritize timeliness and correctness over others. In contrast, tasks like post-event analysis might place higher value on data completeness. From the task-centric perspective, the cross-modal redundancy requirements may vary for different performance goals. For driving on a highway, the buildings as background are redundant data; For localizing and parking, the building data are not as redundant because this task requires more details on the buildings.

\paragraph{Application Layer}

Applications are organically aggregated tasks \cite{hu2023planning}. One task can be mapped to multiple applications, and one application can involve many goals. For example, object detection is a fundamental task across most AD applications \cite{wang2023od}. Recognizing the overall context and sub-tasks can help refine data quality requirements and ensure performance \cite{zheng2024genad, jia2024bench2drive}.

There are different categorizations of autonomous driving applications. A widely adopted application pipeline is a serial design of perception, prediction, planning, and controlling \cite{li2024datacentricAV}. Perception is the detection and interpretation of objects and their surroundings using sensors \cite{han2023collaborative}. Prediction anticipates future movements of detected objects based on current data. Planning refers to deciding an optimal trajectory by the environmental context and the predicted behavior of objects \cite{omeiza2021explanations}. Controlling executes these planned vehicle operations, such as steering, acceleration, and braking. There are also end-to-end approaches integrating the pipelines \cite{Chen2024E2E}. 

\paragraph{Goal Layer}

The goal layer is the highest level within the Vase Framework, providing the overall objectives that applications and tasks aim to achieve, including efficiency, latency, and accuracy. Without this layer, data quality and task optimizations could not get feedback and improve iteratively. Taking pedestrian prediction as an example, the Goal layer defines high-level performances as sufficient accuracy and low response latency. Flagged sensor data informs quality issues in data curation and labeling, and engineers could refine the model to restore speed or precision. 

High efficiency can enhance the driving experience and reduce operational costs and environmental impact in AVs \cite{pan2024impacts}. Latency evaluates if the model’s response time and processing speed satisfy the real-time demands of autonomous driving across data collection, computation, consumption, and communication \cite{gog2017pylot, lin2023latency}. High latency can compromise real-time decision-making capabilities, resulting in delayed reactions to dynamic road conditions and potentially increasing safety risks. High accuracy typically demands high computational power, leading to increased energy consumption and reduced efficiency.

\subsection{Relationships Between Layers}
\label{subsec: relationships}

The relationships between layers are important for the following reasons. First, there are trade-offs in AV design. Higher-resolution sensors can capture smaller or distant objects more reliably, but also increase data processing costs. In addition, enhancing one DQ dimension may come at the cost of another. These trade-offs must be carefully managed to align with overall goals. Second, feedback loops in these layers provide better traceability and interpretability, answering questions such as ``For object detection at night, what are the DQ requirements to achieve an 80\% precision?” Finally, a relational view supports modular iteration and deployment. New sensors, learning algorithms, or application modes can be integrated quickly when their relationships in a task are clear. In sum, analyzing the inter-layer relationships can enhance a coherent, verifiable, and adaptive AD system. The above reasons also clarify that tasks, data, and quality are not isolated elements, but are part of an overall framework.

\begin{figure}
    \centering
    \includegraphics[width=0.9\linewidth]{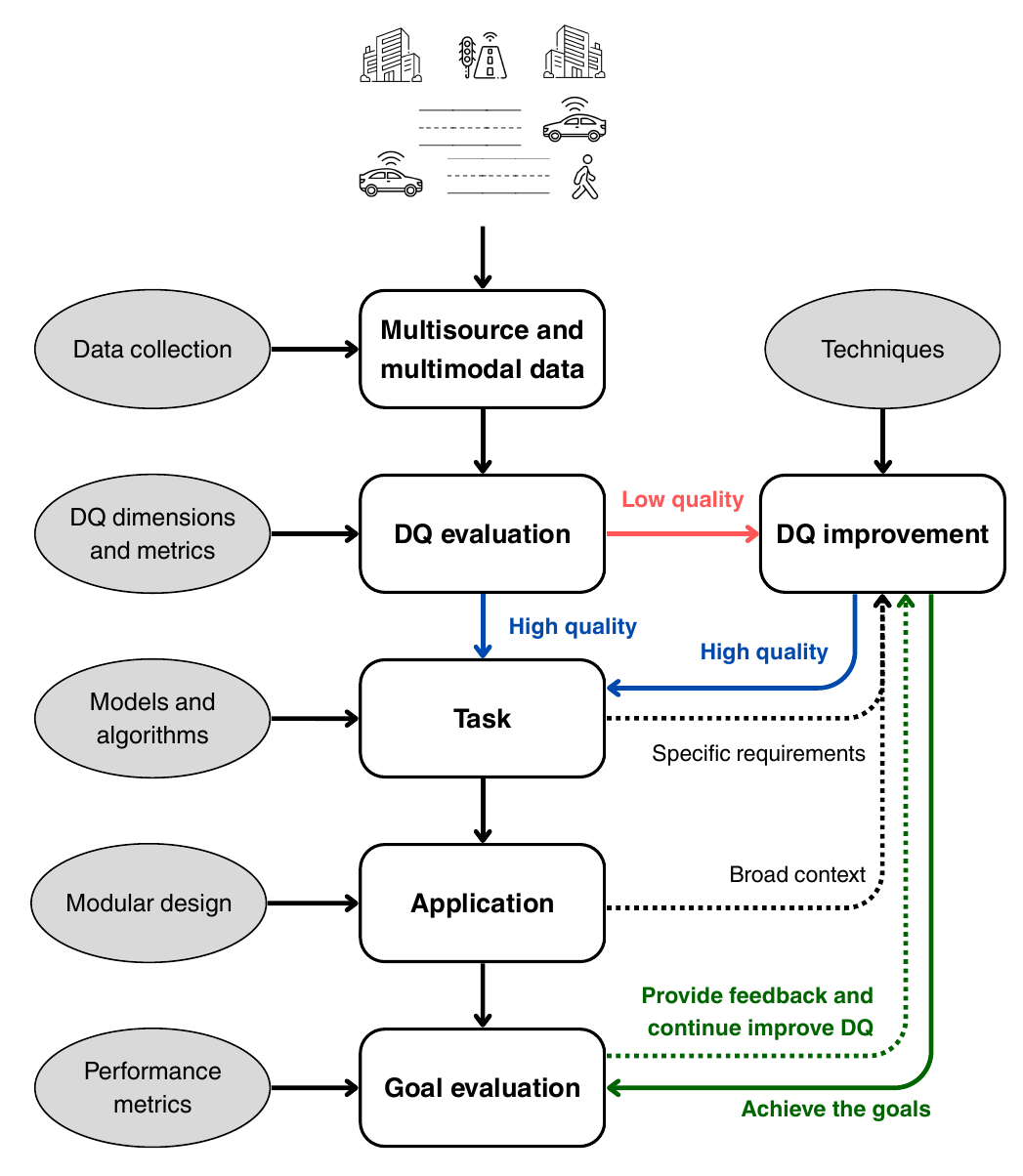}
    \caption{Illustration of the relationships between layers in Vase Framework. The red text and lines indicate low-quality data that requires improvement, and the blue ones represent the high-quality data that can fulfill the task goals. The loops between DQ improvement and task highlight that this data quality framework is task-centric.}
    \label{fig: workflow}
\end{figure}

Fig. \ref{fig: workflow} presents how the framework’s layers interact in loops. Consider an object detection and tracking task within a perception application. Current practice seeks robustness by adding heavier models, stacking sensors, or applying data augmentation strategies \cite{zhang2024igo}. These measures address only parts of the DQ issues and rarely explain how each change influences downstream performance, making the system less adaptable to task requirements.

Under our framework, the workflow changes. Assume the task is performed in a scene with good visibility. During data selection, the pipeline may keep only the image modality and discard LiDAR streams. The subsequent DQ evaluation stage chooses redundancy as its primary dimension because other dimensions, such as correctness, are satisfied under this setting. The evaluation quantifies overlap among camera views and retains only non-redundant information for training. Next, performance metrics validate the data and model. If the reduced redundancy degrades detection accuracy, the loop returns to DQ improvement and relaxes the redundancy thresholds. The flow repeats until the data satisfies the task-specific quality. This process shows how the continuous feedback turns data, data quality, and task performance into mutually reinforcing levers, driving the system toward ever-higher robustness and efficiency.

\section{Case Study}
\label{sec: case}

In this case study, we illustrate the Vase Framework with redundancy evaluation across the multisource camera views and the multimodal camera and LiDAR data for the object detection task in autonomous driving.

\subsection{Dataset}

The nuScenes dataset is a large-scale, multimodal dataset designed for autonomous driving research \cite{caesar2020nuscenes}. It includes 360-degree sensor coverage with data collected from six cameras, five radars, and one LiDAR sensor, along with high-definition maps and detailed object annotations. The cameras are of 12Hz capture frequency. The full dataset consists of 1,000 driving scenes, providing a rich resource for perception tasks such as object detection, tracking, and scene understanding. The nuScenes-mini dataset is a smaller subset containing 10 scenes and 404 frames with the same sensor setting and annotation structure. For this case study, we adopt nuScenes-mini to present our preliminary experiment results.

\subsection{Data Quality: Redundancy}

This case study evaluates the ``Redundancy" dimension to illustrate the relationship between the layers in the framework. Redundant data in AV datasets is related to sensor redundancy. By having multiple sources of similar information, the system can validate data, reducing the impact of errors or noise from individual sensors. When one sensor fails or generates deviated data, the other sensors can still perform \cite{liu2021Multi-sensorRedundancy, qian20223d, he2020sensorRedun}. This enables a more accurate and robust performance \cite{zhang2022sensor, song2019apollocar3d}. In object detection, multiple sensors are set up to obtain comprehensive and reliable environmental information \cite{wang2023od, wang2023scenes}. 

Redundancy also negatively impacts AD system. Firstly, it will decrease the efficiency in computational time and storage \cite{Zhao_2023_ada3d, li2023reducing, Sun2023Task-Driven, xu2024dm3d}. For example, when the weather conditions are good and clear, there is less need to keep high-overlapping information from LiDAR and cameras. The models trained on such data would have little information gain and performance improvement \cite{duong2022active, ju2022extending}. The redundant information includes the temporal correlation between point cloud scans, similar urban environments, and the symmetries in the driving environment caused by driving in opposite directions at the same location \cite{duong2022active, li2023reducing}. As an example, after randomly removing 30\% of the redundant background cloud points, Zhao et al. found only a subtle drop in performance, which proves the necessity and efficiency of researching the redundancy data quality issues in the datasets \cite{Zhao_2023_ada3d}.

Evaluating cross-modal redundancy poses challenges. Each modality has a different informative level and contributes to different tasks \cite{lu2024crossprune}. Reducing redundancy is not simply removing duplicate records, as in previous single-modality data analysis tasks. It requires investigations on the sensor characteristics, task specificity, performance experiments, trade-off evaluation, and so on. 

For the metrics of the redundancy dimension, Li et al. view redundancy as a content data quality issue and divide redundancy into element redundancy and data redundancy \cite{li2023quality}, quantifying how much one or several elements co-occur and repeat the same behavior for longer than the threshold. Duong et al. define redundancy as the similarity between pairs of point clouds resulting from geometric transformations of ego-vehicle movement and environmental changes \cite{duong2022active}. Rao et al. also calculated the similarity between the images and text pairs \cite{rao2020quality}. From an information science perspective, calculating mutual information is also an approach \cite{hadizadeh2024mutual}. 

\subsection{Task Definition}


We take object detection as the task in this case study. It is the process of localizing and categorizing objects of interest \cite{wang2023od}, allowing the system to interpret complex driving environments and make informed decisions based on the presence and movements of objects around the vehicle. High-performance object detection is critical to the safety and efficiency of the whole application.

This task relies on multiple sensors to perceive the environment, each providing complementary data with unique characteristics, including LiDAR, radar, and cameras. Taking multisource and multimodal data as input, it outputs 2D and 3D bounding boxes for the objects. Common detection methods include CNN-based methods such as YOLO \cite{terven2023yolo}, VoxelNet \cite{zhou2018voxelnet}, and PointNet \cite{qi2017pointnet}, transformer models \cite{li2022ViT}, and multimodal fusion like BEVFusion \cite{liu2023bevfusion}. 


The object detection model performance is evaluated using mean average precision (mAP) and recall scores. mAP50 evaluates how accurately bounding boxes align with ground-truth annotations with at least 50\% overlap. Recall evaluates the proportion of actual objects correctly detected or missed.

\subsection{Experiments}

All the experiments were conducted on a PC equipped with one NVIDIA GeForce 4090 RTX GPU and 24 GB of memory. The dataset used for the multi-view camera images is nuScenes-mini. For LiDAR and image detection, we use nuScenes-in-KITTI \cite{NuScenes-in-Kitti} in order to be compatible with the fusion model using KITTI format. The backbone model is YOLOv8, which was trained with a batch size of 16 and 50 epochs. Detailed procedures are introduced in the following sections. 

\subsubsection{Baselines}


Our baseline detection model was evaluated on 1,401 images containing 12,286 object instances across 14 categories of nuScenes. Overall, the detector achieved a box precision of 0.78, a recall of 0.63, and an mAP50 of 0.72. These results serve as the baseline for further investigations on the object detection task using this dataset.

\subsubsection{Redundancy in Multisource and Single-modality data}

This experiment focuses on the multi-view camera images. As shown in Fig. \ref{fig: camera}, the nuScenes camera setup has six pairs of overlapping fields of view (FoV). These overlapping parts indicate areas where cross-camera redundancy may occur. In other words, the front right camera and the back left camera can not capture the same instance simultaneously. Therefore, these six pairs are our research focus. To illustrate our framework by redundancy in this kind of multisource data, we experiment in the following steps:

\begin{figure}
    \centering
    \includegraphics[width=0.75\linewidth]{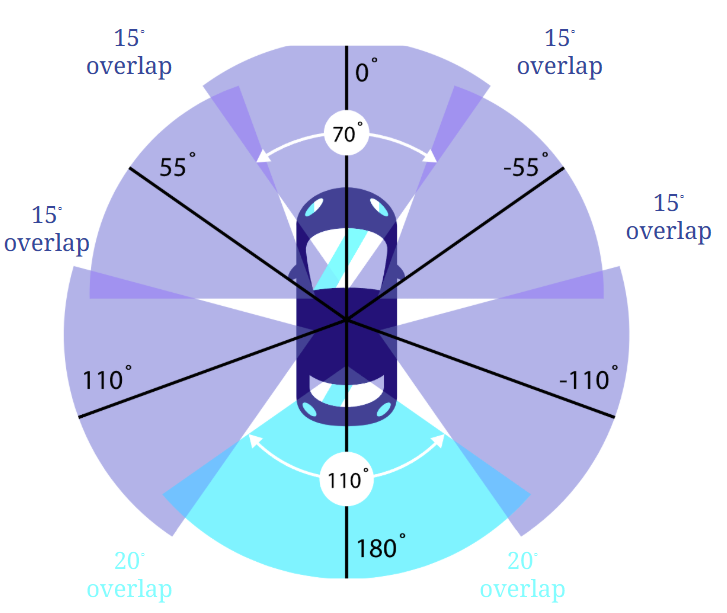}
    \caption{Camera settings and overlapping field of view}
    \label{fig: camera}
\end{figure}

(a) Identify the overlapping FoV based on the nuScenes dataset sensor setting: 

\begin{itemize}
    \item \textbf{Pair 1}: \texttt{CAM\_FRONT} and \texttt{CAM\_FRONT\_RIGHT} overlap by 15° 
    \item \textbf{Pair 2}: \texttt{CAM\_FRONT} and \texttt{CAM\_FRONT\_LEFT} overlap by 15° 
    \item \textbf{Pair 3}: \texttt{CAM\_FRONT\_RIGHT} and \texttt{CAM\_BACK\_RIGHT} overlap by 15°
    \item \textbf{Pair 4}: \texttt{CAM\_FRONT\_LEFT} and \texttt{CAM\_BACK\_LEFT} overlap by 15°
    \item \textbf{Pair 5}: The rear camera \texttt{CAM\_BACK} overlaps with \texttt{CAM\_BACK\_RIGHT} by 20°
    \item \textbf{Pair 6}: The rear camera \texttt{CAM\_BACK} overlaps with \texttt{CAM\_BACK\_LEFT} by 20°
\end{itemize}

\begin{figure}
    \centering
    \includegraphics[width=1.0\linewidth]{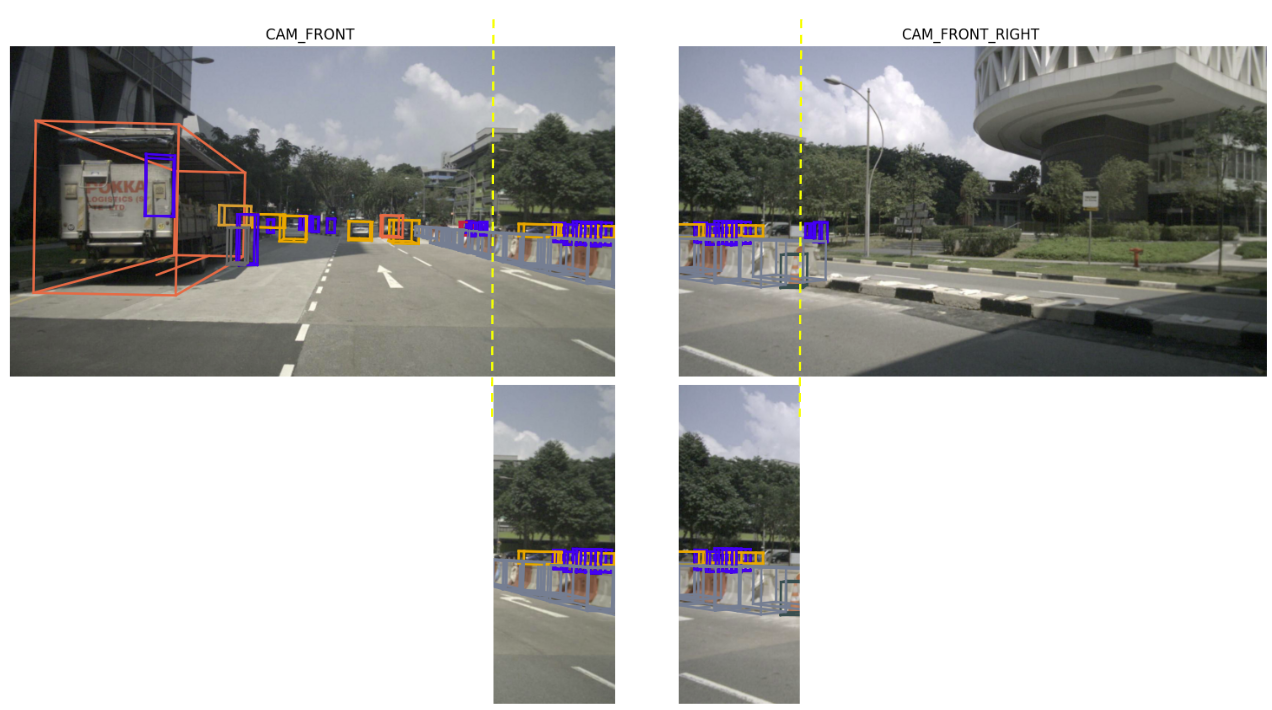}
    \caption{Illustration of cropping based on overlapping angles of the cameras}
    \label{fig: crop}
\end{figure}

(b) Crop images based on overlapped angles: Fig. \ref{fig: crop} illustrates the process of cropping each pair of camera images based on the overlapping FoV. Redundancy occurs in the bottom two cropped images.

(c) Calculate cosine similarity: for the cropped images in the six pairs, we calculate the cosine similarity of each sample and exclude the ones without redundant instances. This provides us with a preliminary understanding of redundancy. 

(d) Create different levels of redundant training datasets to investigate how redundancy affects the inference performance: for each pair of overlapping detections, we compute a \textbf{Bounding Box Completeness Score (BCS)}, which indicates how completely the bounding box (BBox) presents the instance.

Let $\mathrm{BBox}_{\mathrm{full}}$ be the original (uncropped) 2D bounding box area in the image, and let $\mathrm{BBox}_{\mathrm{clipped}}$ be the visible portion after clipping to the image boundaries.  We define:
\begin{equation}
  \mathrm{BCS}(b)
  \;=\;
  \frac{
    \mathrm{Area}\bigl(\mathrm{BBox}_{\mathrm{clipped}}(b)\bigr)
  }{
    \mathrm{Area}\bigl(\mathrm{BBox}_{\mathrm{full}}(b)\bigr)
  }\,,
\end{equation}
where $b$ indexes a candidate box.  Within each redundant group, if
\[
  \max_b \mathrm{BCS}(b)\;-\;\min_b \mathrm{BCS}(b)
  \;>\;
  \tau_{\mathrm{BCS}},
\]
we retain only the box with the higher BCS and discard the lower one; otherwise, we preserve both boxes. As the threshold $\tau_{\mathrm{BCS}}$ increases, fewer boxes are removed, thus retaining more redundancy while still preferring more complete annotations.

(e) Train Yolov8 on each pair of overlapping camera images: from the previous step, we obtain training sets with different levels of redundancy. Next, we train the model using these training datasets and evaluate how removing redundancy affects the inference performance.

\subsubsection{Redundancy in Multisource and Multimodal Data}

Fig. \ref{fig: cross-modal redundancy} shows an example of multimodal data redundancy in the nuScenes dataset: The LiDAR set on top and the front camera capture the same objects. This experiment investigates the redundancy between these two sensors:

\begin{figure}
  \centering
  \includegraphics[width=0.9\linewidth]{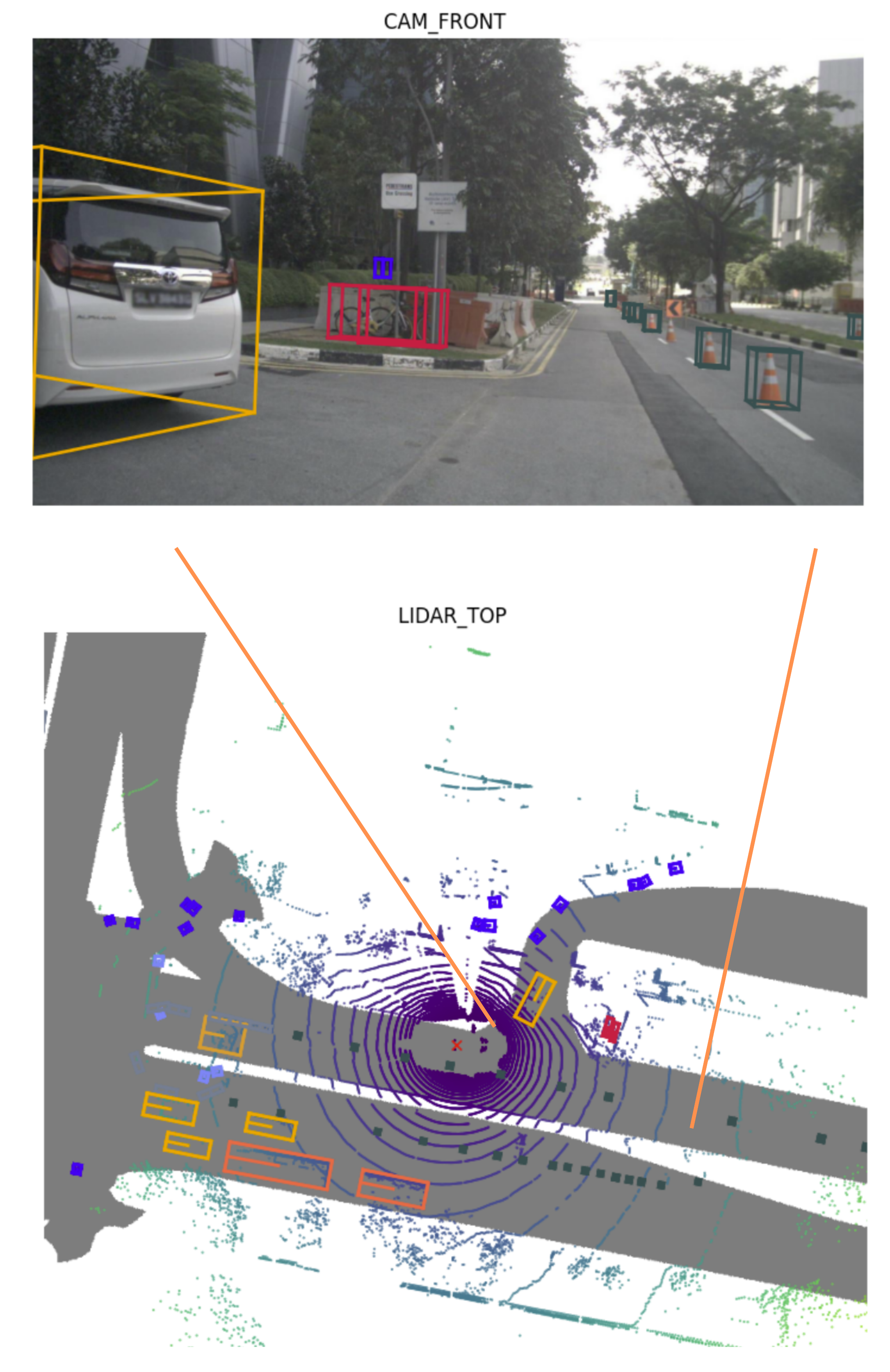}
  \caption{Illustration of the redundancy between the Front Camera and LiDAR}
  \label{fig: cross-modal redundancy}
\end{figure}

(a) Run the full camera-LiDAR fusion detection model \cite{TimKie} to get baseline 3D bounding boxes. Check the overlapping detections between this and the detection using only LiDAR data as redundancy. Let $\mathcal{B}_{\mathrm{LiDAR}}$ be the set of boxes detected using LiDAR only, and $\mathcal{B}_{\mathrm{base}}$ the set detected by the image-LiDAR fusion. The \emph{redundancy ratio} for one frame is
\begin{equation}
  \mathrm{RR}
  \;=\;
  \frac{\bigl|\{\,b\in\mathcal{B}_{\mathrm{base}}\mid\exists\,b'\in\mathcal{B}_{\mathrm{LiDAR}}:\mathrm{IoU}(b,b')\ge\theta\}\bigr|}
       {|\mathcal{B}_{\mathrm{base}}|}\,.
\end{equation}

(b) Compute each box’s 3D centroid in the LiDAR frame, and measure its distance from the ego sensor: 

\[
  \mathbf{c}(b) \;=\; \frac{1}{8}\sum_{i=1}^{8} \mathbf{v}_i
  \quad\text{and}\quad
  d(b) \;=\; \bigl\lVert \mathbf{c}(b)\bigr\rVert_2,
\]
where \(\{\mathbf{v}_i\}\) are the eight corner vertices of \(b\). We then set a single pruning threshold $T_{\mathrm{dist}}$ (meters). All boxes whose centroids lie within this distance are removed:

\begin{equation}
  \mathcal{B}_{\mathrm{pruned}}
  \;=\;
  \bigl\{\,\mathcal{B}_{\mathrm{LiDAR}} \mid d(\mathcal{B}_{\mathrm{LiDAR}}) \;\ge\; T_{\mathrm{dist}}\bigr\}.
\end{equation}

By sweeping \(T_{\mathrm{dist}}\), we could trace how many boxes are pruned and what fraction of true detections is lost, thus choosing a trade-off point that maximally reduces redundancy without harming too much detection performance. This distance threshold is chosen based on statistical results explained in the next section.

(c) Run LiDAR detection again using the partially removed LiDAR data and observe how removing redundancy affects the performance. Let $\mathcal B_{\mathrm{pruned}}$ be the set after pruning. We define the \emph{lost ratio} $\ell$ as the fraction of baseline boxes removed due to removing the redundancy:
\[
  \ell
  \;=\;
  \frac{\bigl|\mathcal B_{\mathrm{base}}\setminus \mathcal B_{\mathrm{pruned}}\bigr|}
       {\bigl|\mathcal B_{\mathrm{base}}\bigr|}
  \;=\;
  1 \;-\;
  \frac{\bigl|\mathcal B_{\mathrm{base}}\cap \mathcal B_{\mathrm{pruned}}\bigr|}
       {\bigl|\mathcal B_{\mathrm{base}}\bigr|}.
\]

\subsubsection{Results}

\begin{figure}
    \centering
    \includegraphics[width=1.0\linewidth]{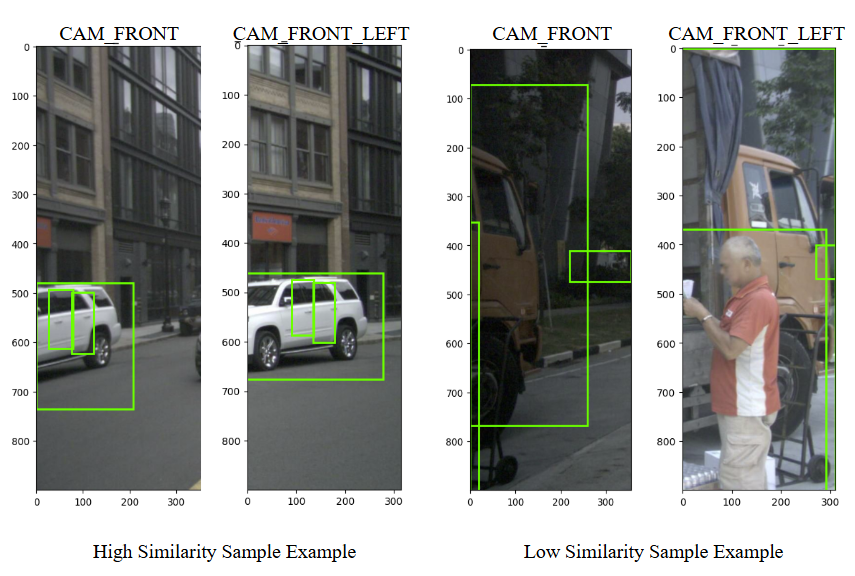}
    \caption{Examples of high and low similarity samples}
    \label{fig: sim}
\end{figure}

For multi-view redundancy evaluation, Fig. \ref{fig: sim} shows an example of high-similarity (left) and low-similarity (right) views in the front and front left cameras. In high-similarity examples (left), the detected objects and background almost provide the same information. By contrast, the low-similarity ones (right) present more complementary detections. 

\begin{figure}
    \centering
    \includegraphics[width=1.0\linewidth]{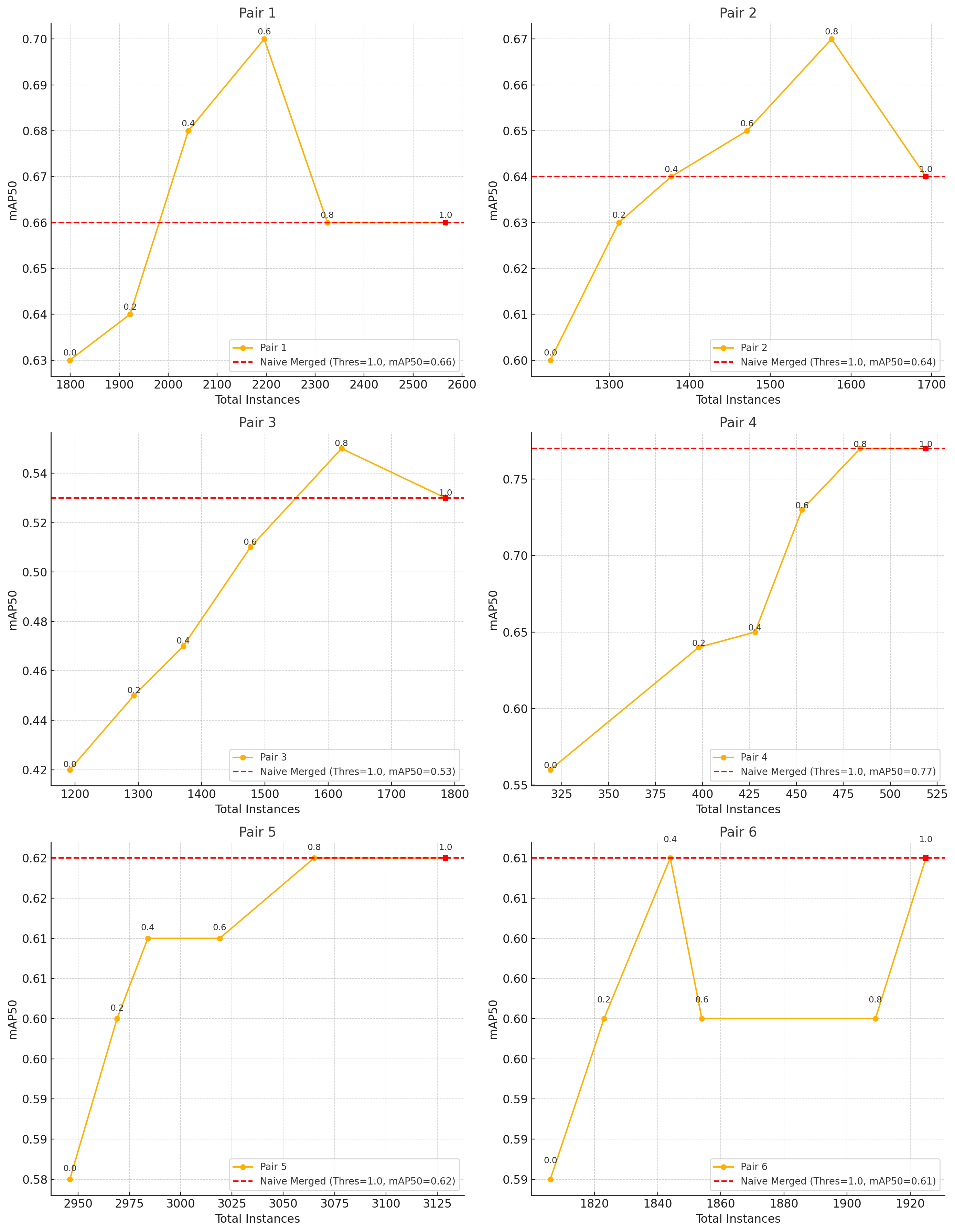}
    \caption{Effects of removing redundant instances in overlapping FoV images on mAP50}
    \label{fig: full pair}
\end{figure}

Fig. \ref{fig: full pair} shows how removing redundancy affects detection performance on six pairs in Fig. \ref{fig: camera}. For each pair, the thresholds are set from 0.0 to 1.0, with a 0.2 interval. As the threshold increases, the instances in the training dataset are kept more, meaning the redundancy is reserved more, till the 1.0 threshold keeps all the instances for training. Interestingly, the trends reveal that a less redundant training dataset can achieve or even surpass the performance level of using the full training dataset.  

\begin{figure}
    \centering
    \includegraphics[width=1.0\linewidth]{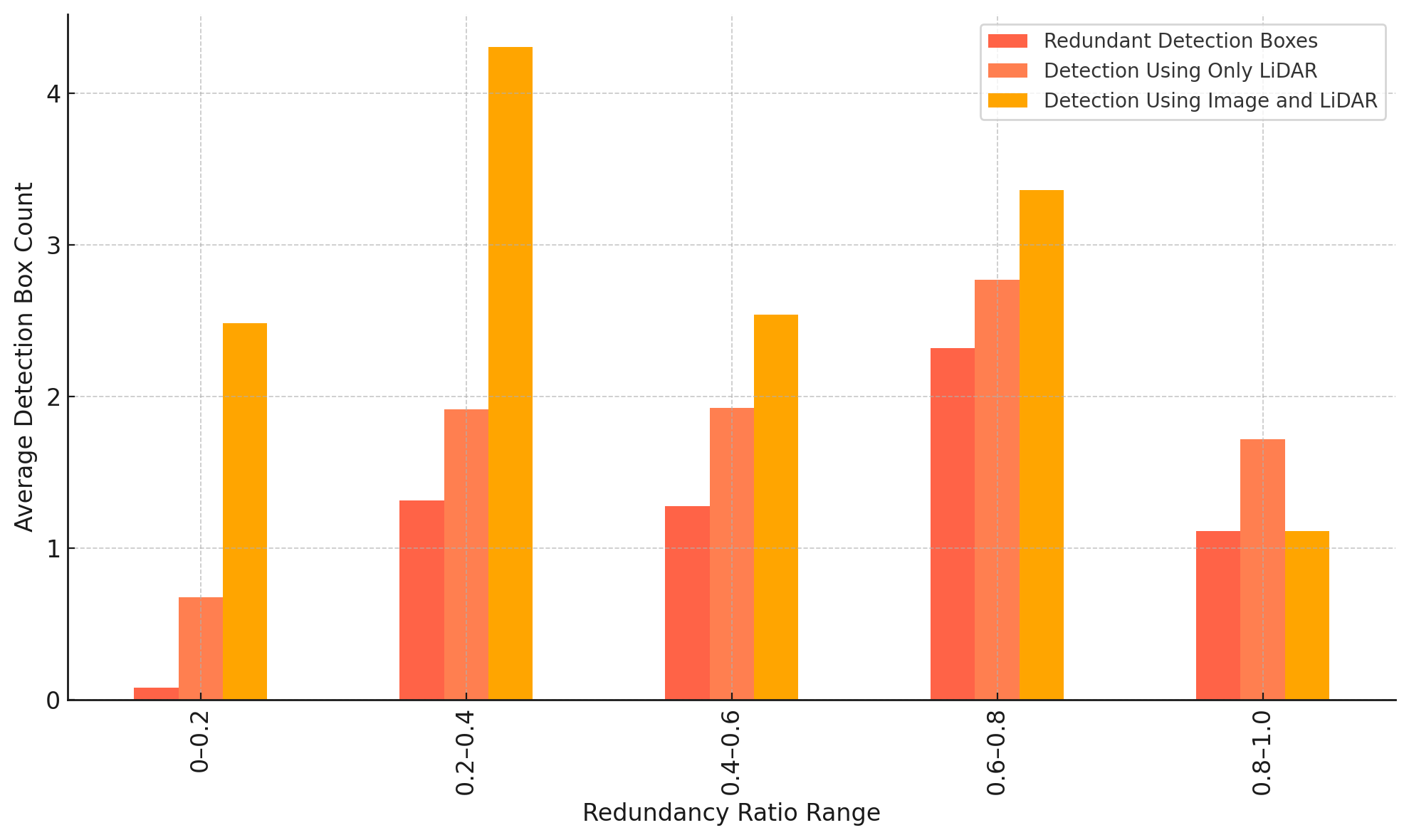}
    \caption{Redundancy in LiDAR and image data}
    \label{fig: lidar-red}
\end{figure}

\begin{figure}
    \centering
    \includegraphics[width=0.85\linewidth]{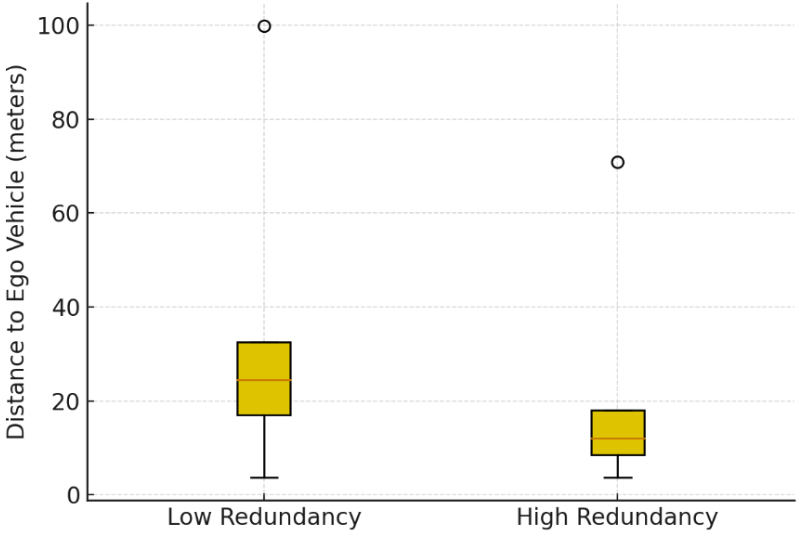}
    \caption{T-test of distance threshold and cross-modal redundancy}
    \label{fig: distance stat}
\end{figure}

\begin{figure}
    \centering
    \includegraphics[width=1\linewidth]{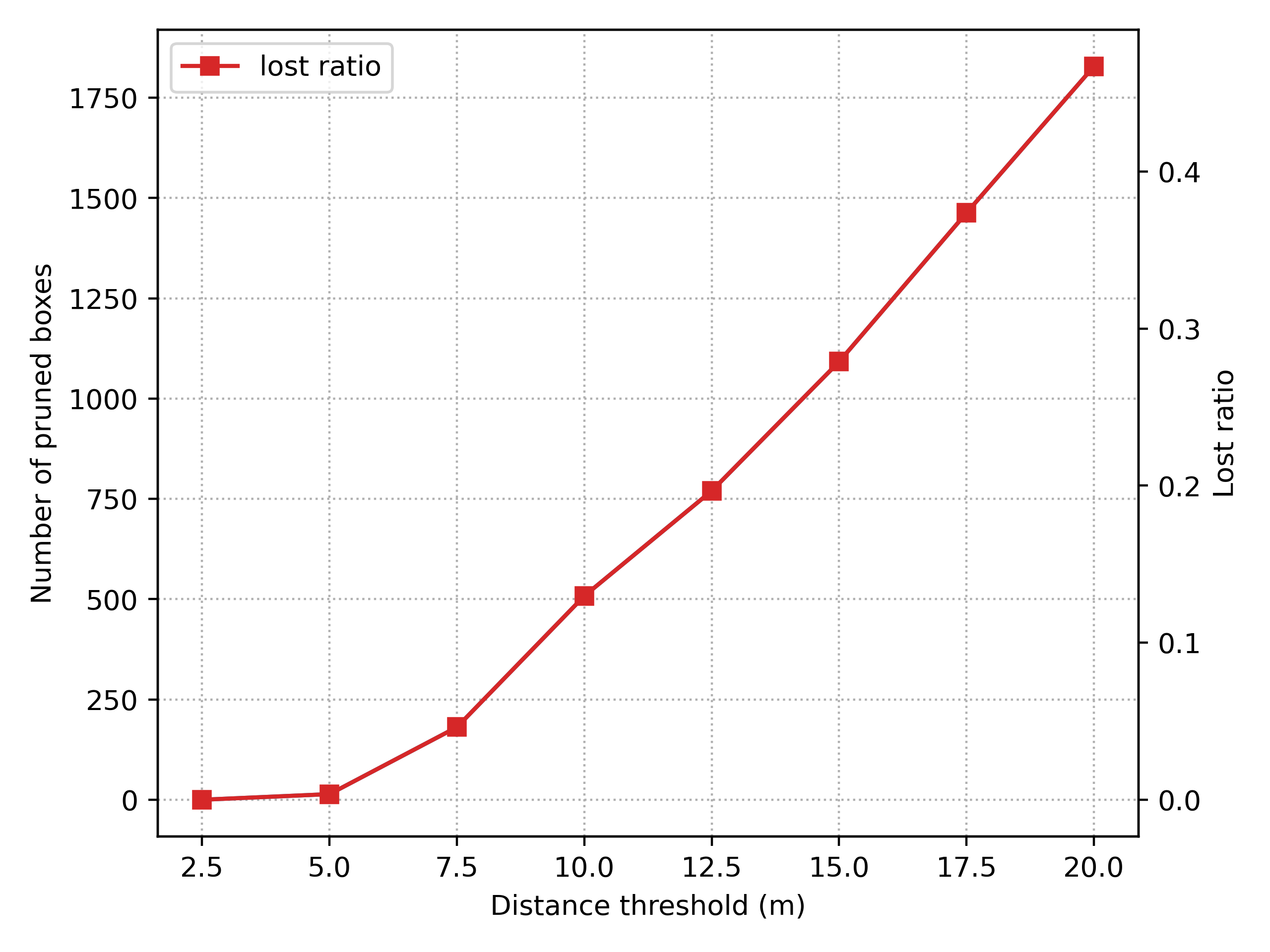}
    \caption{The effects of removing close-range redundant LiDAR data on detection performance}
    \label{fig: distance_pruning_curve}
\end{figure}

For multimodal redundancy evaluation, Fig. \ref{fig: lidar-red} shows the distribution of redundancy across the Lidar and image detection. We further group and compare the high-redundancy and low-redundancy samples. T-test results of distance threshold \(T_{\mathrm{dist}}\) and cross-modal redundancy (p-value=1.17e-76) in Fig. \ref{fig: distance stat} suggest that the high cross-modal redundancy objects remain very close to the ego-vehicle. This supports us in removing the close-range LiDAR data as redundancy. Fig. \ref{fig: distance_pruning_curve} shows that the detections can be viewed as unaffected by choosing a reasonable threshold and removing near-range LiDAR points. It also proves that in many scenes, close-range data is redundant. While removing them could have little impact, the efficiency can be improved due to the decrease in data points to be processed. 

\subsection{Discussion}

First of all, we demonstrate that redundancy exists in (1) multisource and single modality data, which is the multi-view camera images; and (2) multisource and multimodal data, which is the image and LiDAR data, using qualitative illustrations and quantitative statistics in two experiments.

This result is not surprising, as the autonomous driving system requires a certain level of redundancy to cross validate and ensure safety \cite{liu2021Multi-sensorRedundancy, qian20223d, he2020sensorRedun, zhang2022sensor, song2019apollocar3d}. From a data perspective, we want to explore the suitable methods to (1) define and quantify the redundancy at the instance level in the context of object detection, and (2) remove the low-informative yet computation-consuming redundancy, i.e., how to keep the redundancy that helps model performance. 

From the results of the first experiment in Fig. \ref{fig: full pair}, introducing the BCS could effectively reduce redundancy by removing lower-quality duplicates while preserving complementary views when they are of similar quality. Using fewer instances to train the model could perform better than the models trained with more instances because the redundant data does not provide new information. This again proves the significance of data quality rather than quantity. 

Further, this strategy also yields a similar distribution of instances between the sensors as the original unpruned dataset. The original proportion of the number of instances in the front camera-front right camera pair is similar to the proportion in the BCS-guided redundancy-removed dataset. This finding implies that the representation of the instances is an important aspect of data quality improvement.


The evaluation of image-LiDAR redundancy reveals that high redundancy ratios tend to occur for objects located close to the ego-vehicle, where dense LiDAR returns. This suggests that these instances are being detected equally well by both modalities and therefore contribute little new signal when both are included. This finding calls back to the single modality data redundancy that removing redundancy could have very little impact on the performance, while having other benefits, such as reducing the training dataset size.

The BCS in the first experiment also guides us in looking into the relationship between the distance to the ego vehicle and the 3D bounding box size. Zhang et al. find that the distance captured in the point cloud is a crucial factor for achieving a high point cloud quality \cite{zhang2024igo}. Large, well-structured objects are easy for sensors to detect. Ideally, the further the object is located, the smaller the 3D bounding box size is. However, the bounding box quality remains a concern. From qualitative checks, the boxes could not present the object size very well, despite accurately providing the centroid of the location. This was also observed in Kim et al.'s study on label quality \cite{kim2025automatic}. They pointed out that many boxes are labeled as either too small or too big, resulting in low-quality datasets. 

Based on the above, further exploration of the redundancy DQ dimension in AVs will focus on the following perspectives: 

\begin{enumerate}
    \item This paper has initially explored data redundancy using the nuScenes dataset and the YOLO object detection model, demonstrating that partially removing redundant data could potentially improve model performance. We will extend this evaluation across diverse and larger-scale datasets, suitable and SOTA models, and other autonomous driving tasks to thoroughly evaluate redundancy impacts under varied conditions.

    \item Meanwhile, investigating how redundancy levels vary across different driving scenarios, such as dense urban traffic versus highway conditions, as well as diverse lighting and weather environments, could be another direction. These factors likely influence the relationships among multi-view camera data and multimodal sensor inputs, affecting overall redundancy. Therefore, adaptive evaluation and improvement methods are important.
    
    \item Additional modalities such as radar or user-generated data introduce complexity, potentially revealing more redundancy. Integrating these modalities could further refine our conceptual understanding and definition of redundancy. On the other hand, different autonomous driving tasks may impose varying requirements on sensor modalities; adding more modalities does not always equate to improved performance. For example, certain object detection tasks might achieve sufficient accuracy without LiDAR data. In such cases, introducing additional modalities could result in high and negative redundancy.
    
    \item Redundancy's impacts may differ based on the specific task, dataset, or model employed. Our research efforts focus on data-centric analysis. Developing methods to define, detect, and address redundancy in unlabeled datasets, guided by specific task requirements, represents a significant area for future exploration.

\end{enumerate}

Moving forward, there are more important topics and questions in our framework. Here we present our case study as a preliminary foundation and put forward future directions in the following section for discussion.

\section{Open Research Questions and Future Directions}
\label{sec: future}

While our proposed task-centric and data quality (DQ)  framework offers a foundational approach and the case study validates the usefulness of the framework, it also opens up a range of critical but unexplored challenges that merit further investigation. This section outlines key open research questions and promising future directions at the intersection of DQ, task orchestration, and performance-oriented system development in AVs. 

\subsection{Open Research Questions}

\begin{itemize}

    \item \textit{How to develop a taxonomy of task-specific DQ dimensions for multisource and multimodal data in AVs?} A formal taxonomy would help standardize how quality is defined for images, LiDAR, radar, GPS, etc. In addition, how does each DQ type contribute to specific tasks? For example, how do spatial resolution, temporal consistency, and labeling accuracy impact object tracking, while pixel-wise accuracy, label distribution balance, and occlusion influence semantic segmentation? 

    \item \textit{How to design DQ metrics that are task-adaptive and dynamic across AV modules?} Different tasks, such as object detection, lane keeping, trajectory planning, and emergency braking in AVs, may require different sources and modalities of high-quality data. A single static DQ metric may not be sufficient. This question will investigate task-conditioned DQ metrics that dynamically adjust based on the goal and current environmental context. For example, in low-light environments, the DQ metric could weigh LiDAR data higher than image data for perception. The objection detection task in AVs may prioritize visual clarity and spatial resolution, while the trajectory planning task may prioritize temporal continuity. 

    \item \textit{What is the optimal trade-off between data volume, DQ, and system latency in AVs?} As datasets for AVs continue to grow, the cost of storing, processing, and computing for multisource and multimodal data, especially videos and images, will escalate rapidly. Exploit methods for subsampling large datasets and evaluating the most relevant DQ dimensions to a set of downstream multimodal vision and language tasks are critical. This question explores how to mathematically model and optimize trade-offs between DQ (redundancy, completeness, etc), task performance (detection accuracy, etc), and latency (inference time limits, etc). 

    \item \textit{How can the AV systems detect and respond to dynamic DQ changes in real-time?} Sensor degradation due to dirt, fog, rain, occlusion, or hardware failure may compromise data streams. Investigating real-time DQ monitoring mechanisms and autonomous failover strategies will enhance the robustness, adaptability, and safety of AVs. For example, if the front camera is occluded, how can we design an algorithm to guide the system to reweight LiDAR and side cameras for decision-making within a short time?

    \item \textit{How can DQ-aware mechanisms be applied to enhance explainability in AVs?} Explainable artificial intelligence (XAI) in AV aims to understand why a model made a particular decision, which is very important for public trust, regulatory compliance, and system safety. However, current XAI in AVs focuses on the model internals, such as feature attributions, attention weights, or decision trees, while ignoring that DQ could be an equally critical source of error. This question will bridge DQ with XAI by understanding the relationship between the quality of input data and the explanation made in AVs. 

    \item \textit{How can a task-centric DQ evaluation framework be used to support synthetic data generation and augmentation in AVs?} Data synthesis (GANs, simulation) and data augmentation (LLMs) have been widely used in AVs. However, one of the key challenges is to evaluate and select high-quality data for training. The proposed framework can be instrumental in what synthetic data to generate, evaluating the quality of synthetic samples from the perspective of downstream tasks, and helping determine which synthetic data should be integrated into training datasets. 

    \item \textit{How can we integrate feedback from human users into a task-centric DQ framework to enhance data-driven decision-making in AV?} Semi-autonomous driving systems - such as those offering driver-assist features - involve continuous human-machine collaboration. In such systems, the vehicle may perform most tasks, but human drivers still oversee, intervene, or take over under certain conditions. It opens up an interdisciplinary avenue involving human-computer interaction (HCI), cognitive modeling, and data engineering.

\end{itemize}

These questions aim to guide the community toward building more adaptive, explainable, and resilient AVs that respond intelligently to dynamic environments and heterogeneous data streams.

\subsection{Future Directions}

\subsubsection{Unified DQ Standards for Autonomous Driving}

This direction aims to answer the first and second questions in our framework. Integrating the metrics to form a holistic evaluation could provide helpful guidance in DQ research of AV. Further research should explore, define, and establish standardized metrics for critical data quality dimensions, such as correctness, consistency, and completeness, especially in the context of multisource and multimodal data, with the real-time constraints in AV. 

This also requires investigating the loops among the five layers. Data quality, task, and goal form a ``many-to-many" mapping relationship, i.e., many DQ dimensions, many tasks, and many goals. Therefore, beyond our case of redundancy-object detection-precision, which is a  ``one dimension-one task-one goal" loop, further research could conduct extensive studies of ``multi dimension-one task-multi goal", ``multi dimension-multi task-multi goal", etc.

\subsubsection{Quality-aware Data Selection}

Developing an integrated and practical data quality evaluation workflow based on this framework is essential. This overarching approach should specify the methodologies to systematically and iteratively evaluate, improve, and monitor data quality in AV and other important multimodal domains. To select high-quality data and improve low-quality data, future studies could adopt active learning to selectively retain, discard, or request new data based on evolving task demands and data quality profiles.

\subsubsection{Hardware-Aware DQ Optimization on the Edge}

Future work should move beyond software-only fixes and pursue sensor–pipeline co-design. This calls for joint optimization models that balance sensor settings, edge-compute schedules, data processing pipelines, and quality-aware task adaptation. 

\subsubsection{DQ-aware Simulation and Digital Twins for Safety Evaluation}

Current simulators vary weather and traffic but assume ``perfect” data. The next step is to simulate the effects of varying data quality in synthetic environments, such as injecting controlled quality degradation of sensor noise and time-sync drift. Embedding these capabilities in digital-twin platforms would accelerate investigation of the feedback loop, robust testing, certification, and fault injection studies.

\subsubsection{Reinforcement learning agents for DQ-Task-Goal Alignment}

Reinforcement learning (RL) agents could help dynamically adjust data selection, sensor configurations, and computation strategies to optimize task outcomes. By optimizing a reward that aligns DQ metrics with task goals, an RL agent can navigate complex trade-offs. 

\subsubsection{Human-in-the-loop Interfaces for DQ-aware Debugging and Governance}

Data engineers and regulators need to capture how data quality propagates into decisions. Visual analytics dashboards could plot quality metrics alongside task goals, enable data quality-task relationship checks, facilitating simulations and debugging under altered quality thresholds. Advances in this area will facilitate DQ standard design, foster transparency, and build public trust in DQ-aware autonomous systems.

\section{Conclusion}

\label{sec:con}

AV systems intake and output data from diverse sources and modalities, and they need a holistic DQ framework that tailors quality dimensions to each task’s unique demands and ensures data meets required standards at each stage. This underscores the necessity of a framework that aligns data quality with task requirements and overall objectives, bridging data-level quality metrics and application-level goals. Motivated by this, we put forward a five-layered framework and emphasize that the data work in this field should be task-centric and quality-aware. 

The relationships between the five layers guide experiment design. To illustrate, we choose the nuScenes dataset, investigate the data features and explore redundancy as a data quality issue. We then design an object detection task that reflects both task‑level demands and application goals, comparing performance when we leverage multisource images, only LiDAR, or a fusion of both. While images deliver rich semantic cues, LiDAR provides depth and localization information. This again emphasizes the idea that dataset choice should be aligned with the task requirements. By measuring how redundancy removal impacts detection metrics, we demonstrate how our framework could guide data selection, quality evaluation, and task optimization in a structured and goal‑oriented way.

At last, the research goals and open questions point out the following main directions of our study. We will extend this pilot study to demonstrate the framework with our latest findings. 

\section*{Acknowledgment}

They would like to thank Dr. Junhua Ding from the Department of Data Science, Drs. Heng Fan and Yunhe Feng from the Department of Computer Science and Engineering at the University of North Texas for their precious feedback and suggestions.

\bibliographystyle{IEEEtran}
\bibliography{reference}

\end{document}